\pdfoutput=1

\documentclass[11pt]{article}

\usepackage[]{EMNLP2023}

\usepackage{times}
\usepackage{epsfig}
\usepackage{graphicx}
\usepackage{amsmath}
\usepackage{amssymb}
\usepackage{comment}
\usepackage{graphicx}
\usepackage{amsmath}
\usepackage{amssymb}
\usepackage{booktabs}
\usepackage{multirow}
\usepackage{enumitem}
\usepackage{float}
\usepackage{subcaption}
\usepackage{times}
\usepackage{latexsym}
\usepackage{graphicx}
\usepackage{amsmath}
\usepackage{multirow}
\usepackage{tabularx}
\usepackage{inconsolata}
\usepackage{makecell}
\usepackage{booktabs}
\usepackage{enumitem}
\newlist{noitemize}{itemize}{1}
\definecolor{darkpastelgreen}{rgb}{0.01, 0.75, 0.24}
\usepackage[T1]{fontenc}

\usepackage[utf8]{inputenc}

\usepackage{microtype}

\usepackage{inconsolata}
\DeclareSymbolFont{extraup}{U}{zavm}{m}{n}
\DeclareMathSymbol{\varheart}{\mathalpha}{extraup}{86}
\DeclareMathSymbol{\vardiamond}{\mathalpha}{extraup}{87}

\title{Mitigating Fine-Grained Hallucination by Fine-Tuning Large Vision-Language Models with Caption Rewrites}

\author{
~~Lei Wang$^\spadesuit$
~~Jiabang He$^\clubsuit$
~~Shenshen Li$^\clubsuit$
~~Ning Liu$^\vardiamond$
~~Ee-Peng Lim$^\spadesuit$ \\
$^\spadesuit$Singapore Management University, $^\vardiamond$Beijing Forestry University\\ 
$^\spadesuit$University of Electronic Science and Technology of China\\ 
}

\begin{document}
\maketitle
\begin{abstract}
Large language models (LLMs) have shown remarkable performance in natural language processing (NLP) tasks. To comprehend and execute diverse human instructions over image data, instruction-tuned large vision-language models (LVLMs) have been introduced. However, LVLMs may suffer from different types of object hallucinations.  Nevertheless, LVLMs are evaluated for coarse-grained object hallucinations only (i.e., generated objects non-existent in the input image).  The fine-grained object attributes and behaviors non-existent in the image may still be generated but not measured by the current evaluation methods.  In this paper, we thus focus on reducing fine-grained hallucinations of LVLMs. We propose \textit{ReCaption}, a framework that 
consists of two components: rewriting captions using ChatGPT and fine-tuning the instruction-tuned LVLMs on the rewritten captions.
We also propose a fine-grained probing-based evaluation method named \textit{Fine-Grained Object Hallucination Evaluation} (\textit{FGHE}).
Our experiment results demonstrate that ReCaption effectively reduces fine-grained object hallucination for different LVLM options and improves their text generation quality. The code can be found at  \href{https://github.com/Anonymousanoy/FOHE}{https://github.com/Anonymousanoy/FOHE}.
\end{abstract}

\begin{figure*}[!htb]
  \centering
  \includegraphics[width=1.0\linewidth]{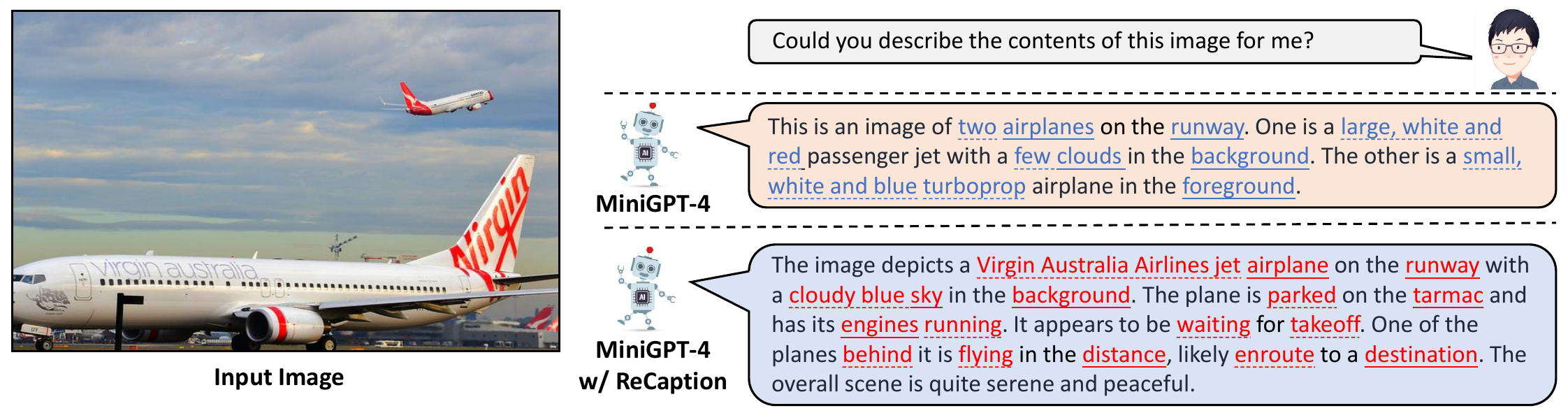}
  \caption{An illustrative example is presented to compare the output of MiniGPT-4 and MinitGPT-4 with ReCaption. The generated caption by MiniGPT-4 contains words in {\color{blue}blue} that are inconsistent with the given image. In contrast, MinitGPT-4 with ReCaption demonstrates a superior ability to generate words in {\color{red}red} that align more closely with the image at a fine-grained level.
  The words marked with an underline represent objects. 
  The words marked with an dotted underline denote attributes and behaviors. 
  }
  \label{fig:intro}
\end{figure*}

\section{Introduction}

Large language models (LLMs), such as GPT-3~\cite{Brown-gpt3-2020} and ChatGPT~\cite{openai-chatgpt-2022}, have demonstrated impressive performance in a wide range of natural language processing (NLP) tasks~\cite{qin-chatgpt-2023}. 
To extend LLMs to comprehend and execute both text-only and multi-modal (i.e., vision + text) instructions, new multi-modal large language models have been introduced, exemplified by GPT-4~\cite{openai-gpt4-2023}. 
Despite the impressive capabilities of GPT-4 in understanding and processing multi-modal information, the underlying mechanisms responsible for these exceptional abilities remain unclear due to its black-box nature.

To shed light on this mystery, recent research endeavors have focused on extending text-only LLMs to comprehend visual inputs by incorporating vision-language models (LVMs) into text-only LLMs.  The resultant model is called the large vision-language model (LVLM).
One LVLM research direction involves using vision modality models to provide textual description for visual information, followed by employing closed-source LLMs, such as ChatGPT, to address multi-modal tasks such as visual QA and image caption generation. 
The LVLM examples using this approach include Visual ChatGPT~\citep{visualchatgpt}, MM-REACT~\citep{mmreact}, and HuggingGPT~\citep{hugginggpt}. 
Nevertheless, this approach requires good alignment across modalities to understand specific multi-modal instructions.

Another alternative approach focuses on instruction-tuned large vision-language models (LVLMs), e.g., LLaVA~\citep{llava}, MiniGPT-4~\citep{zhu2022minigpt4}, mPLUG-Owl~\citep{mplug-owl}, and InstructBLIP~\citep{dai2023instructblip}, which extend language-only open-source LLMs (e.g., FlanT5~\citep{flanT5} and LLaMA~\citep{llama}) to encompass visual reasoning abilities and instruction execution abilities by training LVLMs on text-image pairs and multi-modal instructions. 
Instruction-tuned LVLMs demonstrate outstanding capabilities in solving diverse multi-modal tasks~\citep{Agrawal2019nocaps, schwenk2022aokvqa, scienceqa}.

Despite the success of instruction-tuned LVLMs, these models are prone to hallucination, which compromises both model performance and user experience in real-world use cases~\cite{ji2023survey-text-hallucination}. 
Similar to text-only LLMs~\cite{ji2023survey-text-hallucination, Bang2023evaluate-text-hallucination}, LVLMs may generate text descriptions that include non-existent or inaccurate objects in the target image also known as object hallucination~\cite{biten2022let,rohrbach2018object, ji2023survey-text-hallucination}.

To evaluate object hallucination in instruction-tuned LVLMs, several evaluation methods (e.g., CHAIR~\cite{rohrbach2018object}, POPE~\cite{pope} have been proposed.  However, these methods only focus on coarse-grained object hallucination but not the hallucinated object attributes and behaviors. We call the latter \textit{fine-grained object hallucination}. 
Fine-grained object hallucination refers to the phenomenon wherein LVLMs generate captions that include not only non-existent or erroneous objects but also inaccurate object attributes and behaviors.  

Consider the example in Figure~\ref{fig:intro}. 
This example illustrates that MiniGPT-4, an instruction-tuned LVLM, generates fine-grained hallucinations.
The input image depicts two airplanes: a smaller one in flight and a larger one parked on the runway, accompanied by clouds. Multi-object hallucination occurs when the generated text mistakenly introduces erroneous or irrelevant relations between objects. 
Object attribute hallucination refers to generating incorrect attributes for a particular object. Vanilla MiniGPT-4 generates a description such as ``small, white, and blue turboprop airplane'', even though the color of the turboprop is unknown. Object behavior hallucination pertains to describing incorrect actions for objects. In this example scenario, the smaller airplane is flying, but MiniGPT-4 incorrectly states it is on the runway.

In this paper, we make key contributions to address fine-grained object hallucination.  First, we introduce \textit{ReCaption}, a framework that enables instruction-tuned LVLMs to reduce fine-grained object hallucination by fine-tuning them on additional rewritten captions derived from 
curated high-quality image captions.

ReCaption consists of two components: 1) rewriting captions using ChatGPT and 2) additional training of instruction-tuned LVLM on the rewritten captions.
To develop the first component, we employ a two-stage prompting strategy to guide ChatGPT to generate high-quality image-text pairs. 
In the first stage, we utilize a prompt to tell ChatGPT to extract verbs, nouns, and adjectives from the original input caption.  In the second stage, the extracted verbs, nouns and adjectives are merged into a list which is used in another ChatGPT prompt to generate a rewritten image caption that covers the list of words.  This caption rewriting process is repeated multiple times, creating a diverse collection of captions that still retain the core content of the original caption. At the end of the first stage, we obtain a set of high-quality image-caption pairs.  
The second component of ReCaption performs fine-tuning of the instruction-tuned LVLM using the above set of image-caption pairs to strengthen the model's fine-grained alignment between visual and text modalities.

To better evaluate the proposed method, we introduce a new evaluation method called \textit{Fine-Grained Object Hallucination Evaluation (FGHE)}.
This method assesses how well any LVLM performs in minimizing fine-grained object hallucination by incorporating another evaluation method POPE with the measurement of hallucinated object attributes and behaviors.  Similar to
POPE, FGHE converts object hallucination evaluation into a set of binary classification tasks, prompting instruction-tuned LVLMs with simple Yes-or-No questions about the probed objects (e.g., ``\textit{Is there a car in the image?}'') and about the attributes/behaviors of objects (e.g., ``\textit{Is the man's clothing blue in the picture?}'').
We evaluate the fine-grained hallucination reduction of ReCaption using POHE and FGHE.
Our evaluation results demonstrate that any LVLM adopting the ReCaption framework can effectively reduce fine-grained object hallucination.

\section{Related Work}

\subsection{Large Vision-Language Models}
As text-only LLMs~\citep{Brown-gpt3-2020, llama, openai-chatgpt-2022} show very good results across NLP tasks~\cite{qin-chatgpt-2023}, there are many works extending LLMs to comprehend visual inputs, and  to the development of large vision-language models (LVLMs).
Two primary paradigms have been pursued in this line of research. The first paradigm involves representing visual information through textual descriptions. Closed-source LLMs, such as ChatGPT~\cite{openai-chatgpt-2022}, are used to establish connections between vision modality models, enabling subsequent handling of multimodal tasks. Several notable approaches following this paradigm include Visual ChatGPT~\cite{visualchatgpt}, MM-REACT~\cite{mmreact}, and HuggingGPT~\cite{hugginggpt}.

The second paradigm centers around training LVLMs using vision-language instructions. 
MultiInstruct~\cite{multiinstruct} engages in vision-language instruction tuning, containing various multi-modal tasks involving visual comprehension and reasoning. LLaVA~\cite{llava} employs self-instruct~\cite{self_instruct} to generate instructions and optimize the alignment network's and LLM's model parameters. MiniGPT-4~\cite{zhu2022minigpt4} integrates a visual encoder derived from BLIP-2~\cite{li2022blip} and trains the model using image captions generated by ChatGPT, ensuring that these captions are longer than the training data of BLIP-2. mPLUG-Owl~\cite{mplug-owl} equips LLMs with multimodal abilities through modularized learning, enabling LLMs to support multiple modalities. Lastly, InstructBLIP~\cite{dai2023instructblip} enhances the cross-modal alignment network, empowering the LLM to generate meaningful semantic descriptions for a given image.
By exploring these paradigms, researchers enhance the ability of LLMs to process visual information, thus expanding their applicability and effectiveness in various vision-language tasks.

\subsection{Hallucnation in Large Vison-Language Mdoels}
A recent survey~\cite{ji2023survey-text-hallucination} has thoroughly analyzed studies examining hallucinations in various tasks, including text summarization~\cite{huang2021factual, maynez2020faithfulness, cao2021cliff, tang2021confit, chen2021improving}, dialogue generation~\cite{shuster2021retrieval, wu2021controllable, dziri2021neural}, and vision-language generation~\cite{rohrbach2018object, biten2022let, xiao2021hallucination, Dai2022PlausibleMN, liu2023aligning, gunjal2023detecting, wang2023evaluation, liu2023hallusionbench, yin2023woodpecker, wang2023llm, lee2023volcano}. Specifically, within the domain of vision-language generation, object hallucination can be further classified as intrinsic and external hallucinations. 
Intrinsic hallucinations in vision-language generation refer to generated captions that contain incorrect or non-existent objects in the given image. On the other hand, external hallucinations in vision-language generation refer to generated captions that contain irrelevant objects.
To comprehensively investigate the phenomenon of object hallucination in LVLMs, POPE~\cite{pope} endeavors to conduct an extensive empirical examination of object hallucinations across various LVLMs. 
~\citet{xu2023lvlm} add POPE to the proposed comprehensive evaluation benchmark for evaluating VLVMs.
While previous works mainly focus on hallucinations about the presence or absence of objects, LVLMs may also generate more fine-grained erroneous or incomplete descriptions for target images, specifically regarding incorrect attributes associated with objects.
Therefore, this paper aims to mitigate and evaluate finer-grained hallucinations within LVLMs.


\section{ReCaption Framework}

\begin{figure*}[!htb]
  \centering
  \includegraphics[width=1.0\linewidth]{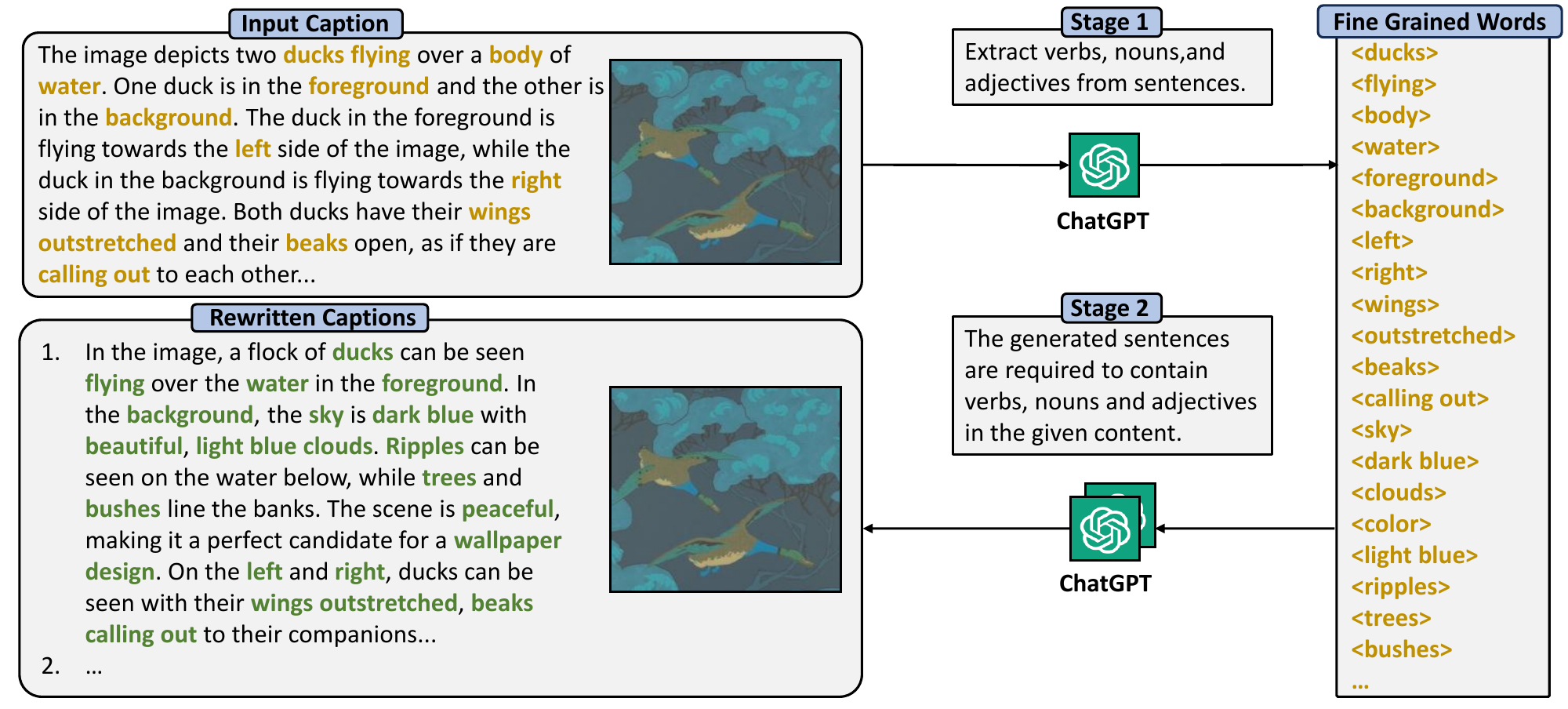}
  \caption{Illustration of rewriting image captions using ChatGPT. Stage 1: Keyword Extraction Prompt (i.e., ``Derive verbs, nouns, and adjectives from sentences''), directs ChatGPT to generate verbs, nouns, and adjectives (highlighted in {\color{brown} brown}) from the original caption. Stage 2: Caption Generation Prompt (i.e., ``The resulting sentences must encompass verbs, nouns, and adjectives aligned with the provided content'') guides ChatGPT to generate a rewritten caption.  By repeating this prompt, multiple rewritten captions will be generated. 
  }
  \label{fig:meth}
\end{figure*}

In this section, we describe the design of our ReCaption framework, highlighting its key components and strategies. ReCaption is LVLM-agnostic and can be used to reduce fine-grained object hallucination of any LVLMs. There are two components in ReCaption, caption rewriting and training of instruction-tuned LVLM. 
The latter is conducted using the rewritten captions. 

\subsection{Caption Rewriting}

The caption rewriting component aims to create good quality captions of training images.  
To generate rewritten image-text pairs, we first randomly select image-text pairs from the cc\_sbu\_align dataset, which is curated by MiniGPT-4~\cite{zhu2022minigpt4}. 
As shown in Figure~\ref{fig:meth}, we use a two-stage prompting strategy on the caption of each selected image
to generate multiple different captions that convey the essential elements of the original caption but with varied details.  We use ChatGPT~\cite{openai-chatgpt-2022} for both stages because it performs well in both stages. We believe other similar LLMs can also be used.
This two-stage prompting aims to preserve the original semantics associated with the corresponding image, which is important for enriching the fine-grained alignment between image and text. 
Note that we provide one demonstration example for each stage to assist LLMs in improving their keyword extraction and caption generation abilities. 
In the following, we elaborate on the proposed two-stage prompting strategy.

\paragraph{Stage 1: Keyword Extraction.}

This stage aims to preserve essential information from the caption corresponding to a specific image. In generating descriptions for an input image, LVLMs may produce inaccurate or irrelevant objects, object attributes, and object behaviors. Since verbs, nouns, and adjectives in captions often denote objects, object attributes, and object behaviors, we devise a prompt to facilitate the extraction of these pertinent words from the original caption. Through this prompting, the extracted keywords preserve the caption's essential information.
Below is the prompt template for the first-stage prompting:
\begin{itemize}[label={}, labelsep=0pt, leftmargin=10pt]
\item \texttt{Extract verbs, nouns, and adjectives from [$X$],}
\end{itemize}
where $X$ denotes the original caption of the input image. 
Figure~\ref{fig:meth} depicts the first-stage prompting directly extracts nouns (e.g., ``ducks''), adjectives (e.g., ``right''), and verbs (e.g., ``flying'') from the caption. The extracted keywords, denoted as $\{x^k_1, x^k_2, ..., x^k_n\}$,  capture essential information in the caption, where $k$ means keywords.

\paragraph{Stage 2: Caption Generation.}

The second stage is to rewrite the original caption, conditioned on the extracted words obtained in stage 1.
The prompt template for second-stage prompting can be denoted as follows:
\begin{itemize}[label={}, labelsep=0pt, leftmargin=10pt]
\item \texttt{The generated sentences are required to contain verbs, nouns and adjectives in the given content: [$x^k_1$, $x^k_2$, ..., $x^k_n$]}.
\end{itemize}
Through prompting ChatGPT, we can obtain a new version of the caption, denoted as $X'$. To keep the characteristic of randomness in the LLM, we use a temperature ratio of $1.0$. Further, we prompt ChatGPT $R$ times 
to generate diverse captions while preserving essential information from the original caption. 
As shown in Figure~\ref{fig:meth}, each rewritten caption has different details while contextually relevant to the original caption. This caption generation prompt basically elicits the imagination and rewriting abilities of ChatGPT.

\subsection{Additional Tuning}

By producing $R$ diverse rewritten captions for each original image caption, we can now enhance the implicit fine-grained alignment between the input images and captions. In our experiments, we set $R=5$. These rewritten image-caption pairs are model-agnostic to language model architecture, thereby enabling their seamless adaptation to various LVLM models, including MiniGPT-4~\cite{zhu2022minigpt4}, LLaVA~\cite{llava}, mPLUG-Owl~\cite{mplug-owl}, and MultiModal-GPT~\cite{Multimodal-gpt}, with minimal changes.
Our later experimental analysis reveals that a small number of rewritten image-text pairs can significantly reduce fine-grained object hallucination and improve caption generation quality. Model tuning using such a small set of pairs entails negligible additional computational cost compared to training of the original LVLM.

The training loss over the images with their rewritten captions can be formulated as follows: 
\begin{equation}
    \mathcal{L} = \sum_{i=1}^{M} \sum_{j=1}^{R} \mathcal{L}_{\mathrm{CE}} (X'_{i,j}, \hat{X}_{i}),
\end{equation}
where $X'_i$ represents a rewritten caption for image-caption pair $i$, $\hat{X}_i$ is generated caption of image-caption pair $i$, $R$ is the total number of rewritten captions for the same image, $M$ denotes the total number of training image-caption examples, and $\mathcal{L}_{\mathrm{CE}}$ is the cross-entropy loss.

\section{Hallucination Evaluation of LVLMs}

To evaluate the effectiveness of ReCaption framework, we use an existing evaluation dataset and method introduced by the POPE work~\cite{pope}.  As POPE only evaluates coarse-grained hallucination, we introduce another dataset with specific evaluation method.  In this section, we will fist introduce POPE dataset and method.  We then introduce our proposed dataset and its evaluation method also known as \textit{Fine-Grained Object Hallucination Evaluation (FGHE)}. 
We leave out the Caption Hallucination Assessment with Image Relevance (CHAIR) metric~\citet{rohrbach2018object} which  suffers from prompt sensitivity and inaccuracies as reported in \cite{pope}. 

\paragraph{POPE dataset and evaluation method.}
In ~\citet{pope}, the Polling-based Object Probing Evaluation (POPE) evaluation method was proposed to improve the evaluation of object hallucination in LVLMs. The basic idea behind POPE is to evaluate hallucination by asking LVLMs simple Yes-or-No questions concerning the probed objects. For instance, a sample question could be: ``Is there a car in the image?''.  By including existent or non-existent object into a question of an input image, the evaluation method obtains a question with ``yes'' or ``no'' answer.  In our experiments, we use popular non-existent objects in the no-questions (i.e., questions with ``no'' answer). Finally, the POPE dataset involves 3000 questions for the captions of 500 images.  By treating a LVLM's answers to the questions as a binary classification task, we obtain the Accuracy, Precision, Recall and F1 scores of the LVLM on this dataset.  The higher the scores, the less hallucination. 
POPE has been adopted by LVLM-eHub, an evaluation benchmark for LVLMs~\cite{xu2023lvlmehub}.
    
\begin{figure}[t]
  \centering
  \includegraphics[width=1.0\linewidth]{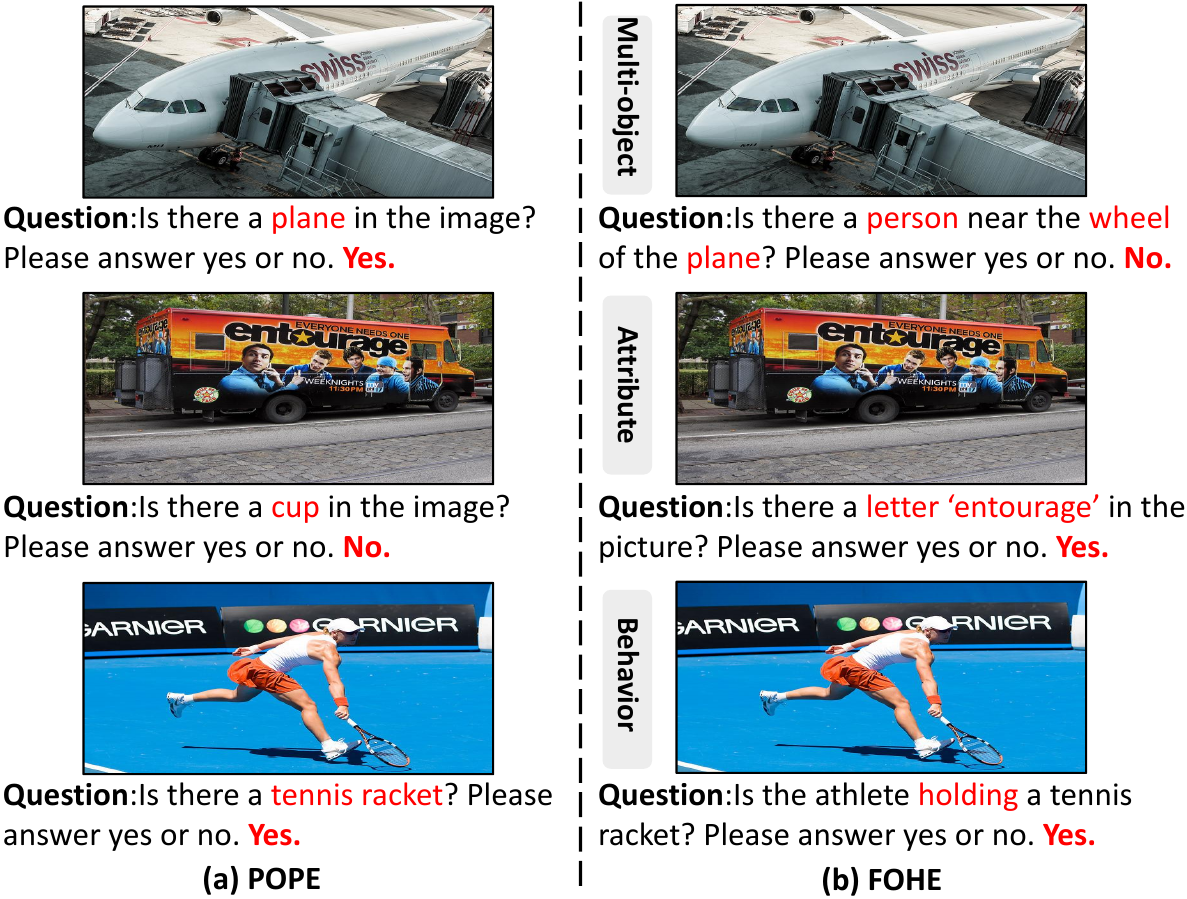}
  \caption{Examples of evaluation data of POPE (a) and FGHE (b).}
  \label{fig:evaluation}
\end{figure}

\paragraph{FGHE dataset and evaluation method.}
\textbf{FGHE} follows the binary classification approach of POPE to evaluate LVLMs. However, unlike POPE, FGHE requires a different set of binary questions to measure fine-grained hallucination.  The FGHE dataset consists of 50 images and 200 binary questions divided into three categories:
(a) \textit{multiple-object} question which verifies the relationships between multiple objects in the image; (b) \textit{attribute} question which verifies an attribute of an object in the image; and (c) \textit{behavior} question which verifies a behavior or an object in the image.  
Figure~\ref{fig:evaluation} presents an illustrative comparison between probing questions in FGHE and POPE.
All the above questions are manually defined by human annotators on a subset of 50 images from the validation set of MSCOCO dataset.
As shown in \label{tab:stat}, the FGHE dataset consists of 100 yes-questions and 100 no-questions.  Among the yes-questions, 47, 45 and 8 are multi-object, attribute and behavior questions. Among the no-questions, 51, 42 and 7 are multi-object, attribute and behavior questions. 
Table~\ref{tab:stat} only displays few behavior questions as some of them are counted towards multiple objects.
Similar to POPE, we finally employ Accuracy, Precision, Recall, and F1 score of all questions as the evaluation metrics.

\begin{table}[]
\centering
\small
\resizebox{0.46\textwidth}{!}{
\begin{tabular}{l c c c c}
\toprule
Type & Questions & Multi-Object &  Attribute & Behavior \\
\midrule
Yes & 100 & 47 & 45 & 8\\
No & 100 & 51 & 42 & 7\\
\midrule
Total & 200 & 98 & 87 & 15\\
\bottomrule
\end{tabular}
}
\caption{Statistics of FGHE Evaluation Dataset.}
\label{tab:stat}
\end{table}


\section{Experiment}

\begin{table*}[htbp] 
\small
\centering
\begin{tabular}{l|cccc|cccc}
\toprule
 \multirow{2}{*}{Model} & \multicolumn{4}{c|}{POPE} &\multicolumn{4}{c}{FGHE} \\
  & Accuracy & Precision & Recall & F1 Score & Accuracy & Precision & Recall & F1 Score \\
\midrule
 mPLUG-Owl       & 59.37   & 55.51   & 94.34  & 69.89  &  55.56 & 57.52 & 84.42 & 68.42  \\
 \,\,\,\, w/ ReCaption       &\bf 61.79   &\bf 57.11   &\bf 94.61  &\bf71.22  &\bf  62.22 &\bf 62.26 &\bf 85.71 &\bf 72.13  \\
 &\textcolor{blue}{$(+2.42)$} &\textcolor{blue}{$(+1.60)$}  &\textcolor{blue}{$(+0.27)$}
 &\textcolor{blue}{$(+1.33)$} &\textcolor{blue}{$(+6.66)$}  &\textcolor{blue}{$(+4.74)$}
 &\textcolor{blue}{$(+1.29)$} 
 &\textcolor{blue}{$(+3.71)$} \\
 \midrule

     LLaVA          & 50.00   & 50.00   &\bf100.00  &66.67  &  56.83 & 56.52 &\bf 100.00 & 72.22  \\
     
     \,\,\,\, w/ ReCaption       &\bf 61.87   &\bf 57.33   & 92.80  &\bf70.88  &\bf  67.63 &\bf 65.42 & 89.74 &\bf 75.68  \\
     &\textcolor{blue}{$(+11.87)$} &\textcolor{blue}{$(+7.33)$}  &\textcolor{red}{$(-7.80)$}
 &\textcolor{blue}{$(+4.21)$} &\textcolor{blue}{$(+10.80)$}  &\textcolor{blue}{$(+8.90)$}
 &\textcolor{red}{$(-10.26)$} 
 &\textcolor{blue}{$(+3.46)$} \\
      \midrule
     
    MultiModal-GPT & 50.00   & 50.00   &\bf 100.00  &66.67  &57.56 &56.93 &\bf100.00 &72.56  \\
    
    \,\,\,\, w/ ReCaption       &\bf 62.40   &\bf 57.60   & 93.93  &\bf71.41  &\bf  75.83 &\bf 72.31 & 83.73 &\bf 77.60  \\
    &\textcolor{blue}{$(+12.40)$} &\textcolor{blue}{$(+7.60)$}  &\textcolor{red}{$(-6.07)$}
 &\textcolor{blue}{$(+4.74)$} &\textcolor{blue}{$(+18.27)$}  &\textcolor{blue}{$(+15.38)$}
 &\textcolor{red}{$(-16.27)$} 
 &\textcolor{blue}{$(+5.04)$} \\
     \midrule
    
     MiniGPT-4       &54.23    &58.24    & 31.97  &41.29  &53.01    &59.62    & 63.27  &61.39  \\

     \,\,\,\, w/ ReCaption       &\bf 57.69   &\bf  62.54   & \bf 39.67  &\bf 48.55  &\bf  60.24   &\bf  62.12   & \bf 83.67  &\bf 71.30  \\
     &\textcolor{blue}{$(+3.46)$} &\textcolor{blue}{$(+4.30)$}  &\textcolor{blue}{$(+7.70)$}
 &\textcolor{blue}{$(+7.26)$} &\textcolor{blue}{$(+7.23)$}  &\textcolor{blue}{$(+2.50)$}
 &\textcolor{blue}{$(+20.40)$} 
 &\textcolor{blue}{$(+9.91)$} \\
    
\bottomrule
\end{tabular}
\caption{
Results of LVLMs w/o ReCaption and LVLMs w/ ReCaption on POPE and FGHE datasets. The best results are denoted in bold.
}
\label{tab:POPE}
\end{table*}

\subsection{Experimental Setup}

\paragraph{Model Settings.}
Since the proposed ReCaption is model-agnostic to instruction-tuned LVLMs, we can enrich any LVLMs with ReCaption. In this paper, we choose four open-sourced representative instruction-tuned LVLMs for evaluation: mPLUG-Owl~\cite{mplug-owl}, LLaVA~\cite{llava}, Multimodal-GPT~\cite{Multimodal-gpt}, and MiniGPT-4~\cite{zhu2022minigpt4}.

\paragraph{Implementation Details.}
A total of $500$ image-caption pairs are randomly selected from a high-quality well-aligned image-text pair dataset named cc\_sbu\_align, which is curated by MiniGPT-4~\cite{zhu2022minigpt4}, for our study. To increase the diversity of the data, we generated $5$ rewritten versions for each original caption. We employed the AdamW optimizer with a beta value of ($0.9$, $0.98$) for optimization purposes. The learning rate and weight decay were set to $0.0001$ and $0.1$, respectively. During the training process, we initiated a warm-up phase consisting of $2,000$ steps, after which we applied the cosine schedule to decay the learning rate. As for the input image, it was randomly resized to dimensions $224 \times 224$. 
We additionally fine-tune LVLMs $20$ epochs with the batch size 256, and the learning rate is set to $0.00002$. 

\subsection{Main Results}

\noindent\textbf{LVLMs with ReCaption have reduced hallucination.}
Overall, a remarkable improvement in performance can be observed across various LVLMs and evaluation metrics when the LVLM is used with ReCaption, as compared to the original LVLM.
For instance, Mini-GPT4 with ReCaption demonstrates a significant improvement of F1 over Mini-GPT4 without ReCaption (7\% improvement on POPE and 10\% on FGHE). Among the LVLMs, mPLUG-Owl with ReCaption enjoys the least improvement in F1 score (with 1.32\% improvement on POPE but more substantial 3.71\% improvement on FGHE).

The results above demonstrate that our proposed ReCaption framework enhances the generation quality of LVLMs. It effectively reduce both coarse-grained and fine-grained hallucinations.  The improvement on fine-grained hallucinations can be attributed to a strong alignment between images and text descriptions at the fine-grained level. Additionally, the proposed ReCaption approach is model-agnostic, allowing for effortless integration as a plug-and-play component during the training of LVLMs.

\noindent\textbf{LVLMs with ReCaption reduce over-confidence.}
As shown in Table~\ref{tab:POPE}, based on the Recall of POPE and FGHE, it is evident that mPLUG-Owl, LLaVA, and MultiModel-GPT show a strong inclination to respond with the affirmative answer ``Yes''. For instance, both LLaVA and MultiModel-GPT provide ``Yes'' responses to most questions. The mPLUG-Owl also achieve a high recall rate of 94.34\% on POPE and 84.42\% on FGHE. It suggests that certain LVLMs exhibit a high degree of over-confidence and struggle to accurately identify objects, object attributes, and object behaviors in the given images.  With ReCaption, LLaVA and MultiModal-GPT reduces their over-confidence.

\noindent\textbf{FGHE serves as a different hallucination evaluation compared to POPE.}
Table~\ref{tab:POPE} reveals different performances of LVLMs evaluated on POPE and FGHE, suggesting that LVLMs possess varying degrees of coarse-grained and fine-grained object hallucinations. For example, the mPLUG-Owl model attains an F1 score of 69.89\% on POPE but only 68.42\% on FGHE, indicating that mPLUG-Owl may excel in identifying objects rather than attributes or behaviors within a given image. Among the LVLMs, MultiModal-GPT performs the best for both coarse-grained and fine-grained object hallucinations. 
\begin{table}[]
\centering
\small
\resizebox{0.48\textwidth}{!}{
\begin{tabular}{l | c c c }
\toprule
  Method  & Multi-Object  & Attribute & Behavior   \\
\midrule
mPLUG-Owl & 72.36 & 68.14  & 61.59\\
\,\, w/ ReCaption & \bf74.23 & \bf71.55  & \bf67.90\\
&\textcolor{blue}{$(+1.87)$} &\textcolor{blue}{$(+3.41)$}  &\textcolor{blue}{$(+6.31)$} \\
\midrule
LLaVA & 74.26 & 71.49  & 66.45\\
\,\, w/ ReCaption &\bf 76.70 & \bf75.18  & \bf71.82\\
&\textcolor{blue}{$(+2.44)$} &\textcolor{blue}{$(+3.69)$}  &\textcolor{blue}{$(+5.37)$} \\
\midrule
MultiModal-GPT & 74.82 & 70.56  & 68.13\\
\,\, w/ ReCaption & \bf76.84 & \bf78.92  & \bf74.22\\
&\textcolor{blue}{$(+2.02)$} &\textcolor{blue}{$(+8.36)$}  &\textcolor{blue}{$(+6.09)$} \\
\midrule
MiniGPT-4 & 63.30 & 60.16  & 56.72\\
\,\, w/ ReCaption & \bf 71.02 & \bf72.16  & \bf67.93\\
&\textcolor{blue}{$(+7.72)$} &\textcolor{blue}{$(+12.00)$}  &\textcolor{blue}{$(+11.21)$} \\
\bottomrule
\end{tabular}
}
\caption{Evaluation over different hallucination categories in terms of F1 score of FGHE.} 
\label{tab:stat}
\end{table}

\begin{figure}[t]
    \centering
    \begin{subfigure}{0.236\textwidth}
        \centering
        \includegraphics[width=1.0\linewidth]
        {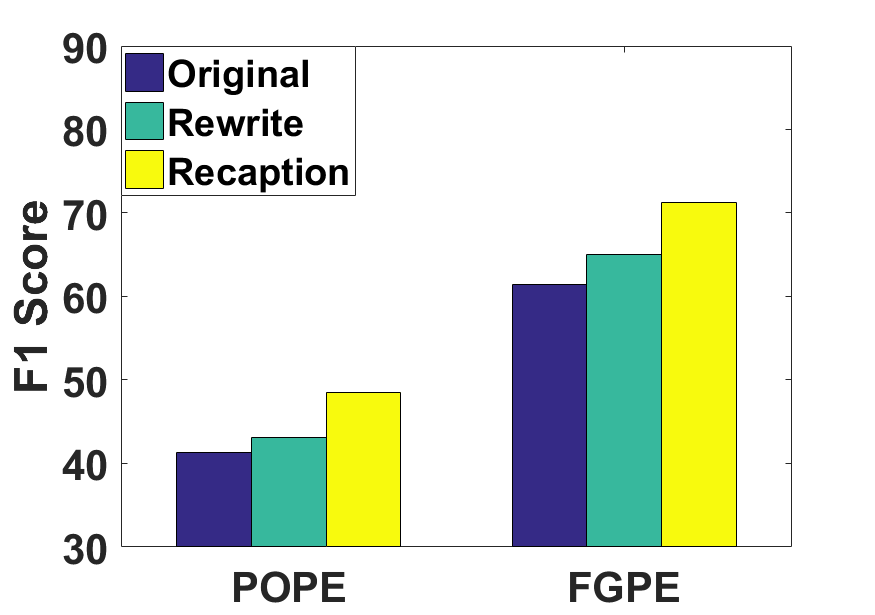}
        \caption{}
        \label{fig:bar_rewrite_1}
    \end{subfigure}
    \begin{subfigure}{0.236\textwidth}
        \centering
        \includegraphics[width=1.0\textwidth]
        {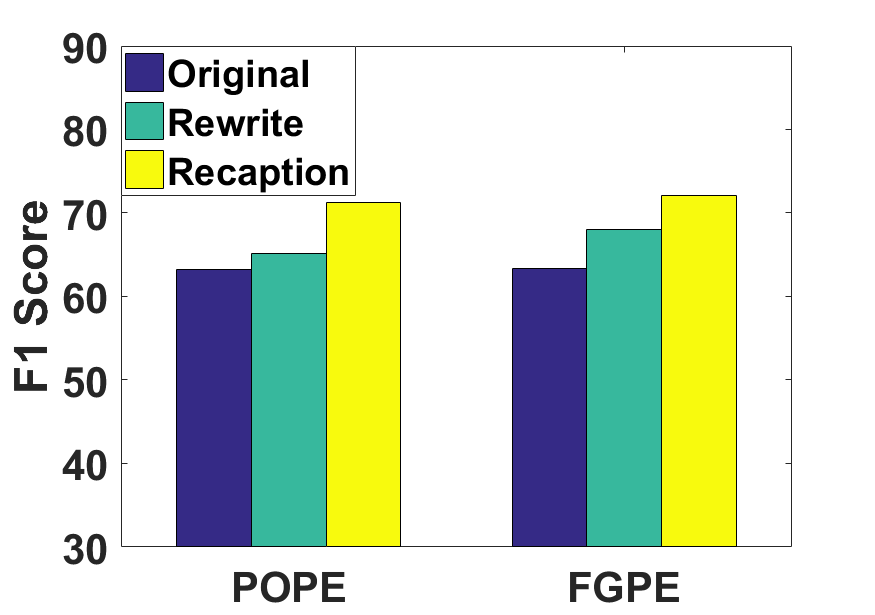}
        \caption{}
        \label{fig:bar_rewrite_2}
    \end{subfigure}
    \caption{Performance comparison of MiniGPT-4 training with different caption rewriting strategies. Original means MiniGPT-4 without rewriting. Rewrite denotes MiniGPT-4 with a simple rewriting prompting (e.g, ``Rewrite the following image description''). ReCaption means MiniGPT-4 with our proposed strategy. }
    \label{fig:rewrite_result}
\end{figure}

\begin{figure}[t]
    \centering
    \begin{subfigure}{0.236\textwidth}
        \centering
        \includegraphics[width=1.0\linewidth]
        {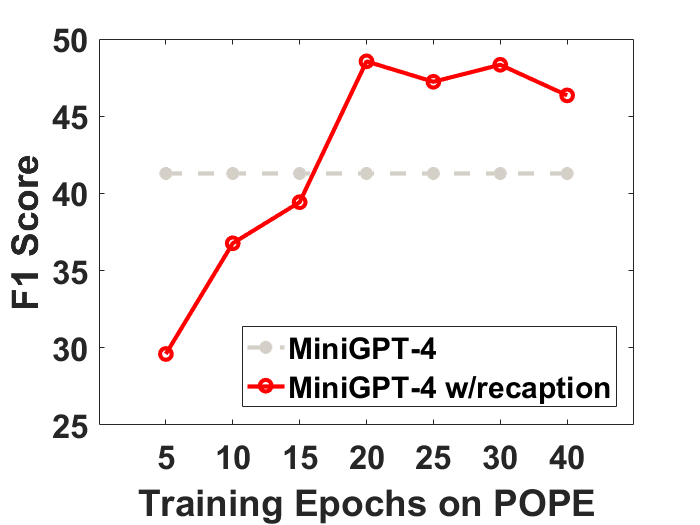}
        \caption{}
        \label{fig:curve_epochs_1}
    \end{subfigure}
    \begin{subfigure}{0.236\textwidth}
        \centering
        \includegraphics[width=1.0\textwidth]
        {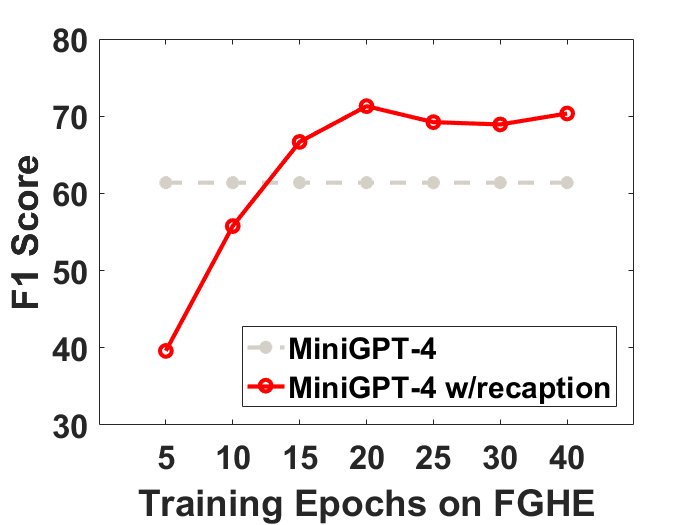}
        \caption{}
        \label{fig:curve_epochs_2}
    \end{subfigure}
    \caption{F1 Score Curves for MiniGPT-4 with and without ReCaption over training epochs. The evaluation datasets used are POPE (a) and FGHE (b).}
    \label{fig:curve_result}
\end{figure}

\begin{figure}[t]
    \centering
    \begin{subfigure}{0.236\textwidth}
        \centering
        \includegraphics[width=1.0\linewidth]
        {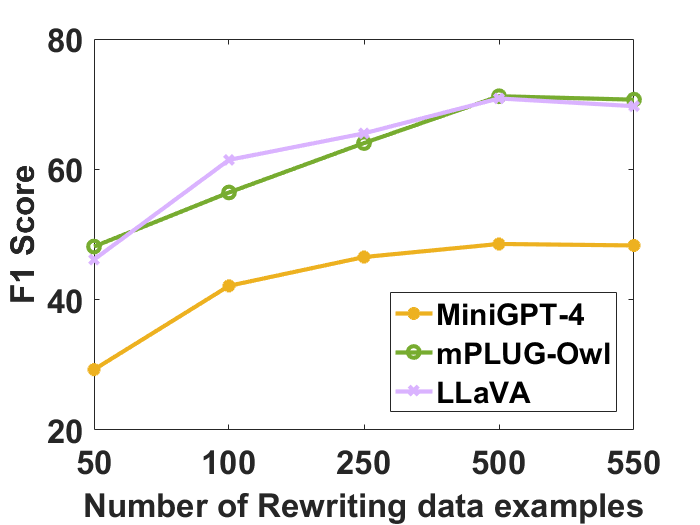}
        \caption{}
        \label{rewriter_samples}
    \end{subfigure}
    \begin{subfigure}{0.236\textwidth}
        \centering
        \includegraphics[width=1.0\textwidth]
        {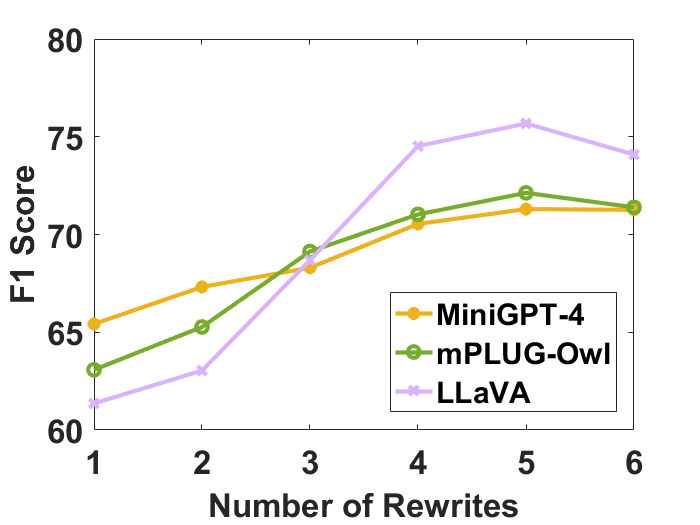}
        \caption{}
        \label{fig:rewriter}
    \end{subfigure}
    \caption{F1 Score Curves of three LVLMs with our ReCaption as the number of images used for rewriting and the number of rewrites per image increase. We evaluate hallucination degree of models using FGHE.}
    \label{fig:curve_result_rewrite}
\end{figure}

\subsection{Further Analysis}

\paragraph{Break-down Study of Fine-Grained Object Hallucination.}
FGHE includes three types of binary questions, namely multiple objects, object attributes, and object behaviors. Table~\ref{tab:POPE} depicts the F1 Scores of FGHE, combining the three types of questions. We further examine the hallucination organized by the three types of questions using the F1 scores.

Table~\ref{tab:stat} presents the results, demonstrating that adding ReCaption to any LVLM reduces fine-grained hallucinations measured by the F1 scores computed over three question categories. This finding suggests that ReCaption is effective for reducing fine-grained hallucination generation by enriching alignment between images and texts through additional training with rewritten captions.
Furthermore, ReCaption is also more effective in reducing hallucination in object attributes and behaviors than in multiple objects. One possible reason is that the rewritten captions may contain more verb and adjective keywords, thereby reducing hallucination, especially in object attributes and behaviors.

\paragraph{Varying Rewriting Strategies.} 
Figure~\ref{fig:rewrite_result} depicts a comparison between our proposed two-stage caption rewriting strategy and a simple rewriting method (called Rewrite) employing ChatGPT~\cite{fan2023improving}. 
We select MiniGPT-4 as the model of choice since the data used for rewriting is based on image-text pairs collected by MiniGPT-4.
The latter uses the prompt ``Rewrite the following image description'' to generate a new caption without any constraint. The performance of this simple rewriting technique exhibits some improvements compared to the original LVLMs without rewrites. However, this improvement is relatively minor. In contrast, our proposed two-stage rewriting prompting approach yields substantial improvements, highlighting multiple diverse rewritten captions with the same keywords to guide the LVLM in establishing a more robust alignment between images and captions.

\paragraph{Number of Training Epochs.} 
We examine the impact of the number of additional training epochs on hallucination reduction of LVLMs with ReCaption. Figure~\ref{fig:curve_result} presents the F1 Score curves in relation to the training epochs for MiniGPT-4 with ReCaption on POPE and FGHE. The figure illustrates that the F1 Score of MiniGPT-4 with ReCaption increases as the number of training epochs increases. This suggests that conducting further training using this limited number of rewritten data pairs can enhance the fine-grained alignment between images and text, consequently reducing the generation of hallucinations by LVLMs.

\paragraph{Number of Images Used for Rewriting.}  
Figure ~\ref{rewriter_samples} illustrates the impact of the number of images utilized for rewriting in additional training of LVLMs, including MiniGPT-4, mPLUG-Owl, and LLaVA. All are evaluated in terms of F1 Score of FGHE. In the figure, a value of 50 indicates LVLMs trained with ReCaption incorporating 50 distinct images and their respective rewritten captions. The remaining hyperparameters, such as the number of rewritten captions per image, remain unchanged and fixed. The results strongly indicate that our proposed ReCaption consistently enhances the performance of all LVLMs as more images are employed for rewriting.

\paragraph{Number of Rewrites per Image.} In Figure~\ref{fig:rewriter}, we observe the effect of the number of rewritten captions per image on LVLMs, such as MiniGPT-4, mPLUG-Owl, and LLaVA. We use the F1 Score of FGHE to evaluate them. Increasing the number of rewritten captions per image is expected to enhance the alignment between input images and output captions. This is because more output captions containing the same keywords are utilized for training the mapping from image to text. The findings strongly support that our proposed ReCaption consistently improves the hallucination reduction ability of all LVLMs as more rewritten captions per image are used during training.

\paragraph{Case Study.}
In the appendix, we show more examples of image caption generation using four vanilla LVLMs and these LVLMs with our ReCaption. Overall, adding ReCaption to LVLMs yields superior quality captions with fewer hallucinated objects, object attributes, and object behaviors. Conversely, based on the provided examples, Mini-GPT4 performs poorly in generating accurate descriptions of the given image, while MultiModal-GPT and LLaVA are prone to generating irrelevant content. Despite vanilla mPlug-Owl displaying a relatively higher-quality caption generation compared to the other three LVLMs, ReCaption still allows it to generate more accurate image captions.




\section{Conclusion}
In this paper, we aim to address the issue of fine-grained object hallucinations in instruction-tuned large vision-language models (LVLMs). We introduced a framework called ReCaption, which comprises two components: caption rewriting using ChatGPT and fine-tuning of LVLMs based on the rewritten captions. To evaluate the effectiveness of our approach, we proposed a fine-grained probing-based evaluation dataset and method called Fine-Grained Object Hallucination Evaluation (FGHE). Our experimental results demonstrate that ReCaption can effectively mitigate fine-grained object hallucinations across various LVLMs, thereby enhancing text generation quality. Future work could focus on refining and expanding the evaluation metrics to support more comprehensive evaluation of LVLM performance in hallucination reduction.  One should also explore more effective rewriting techniques to enrich alignment between unnatural images, such as invoices and cartoon pictures, and longer output text.

\section{Limitations}
Although ReCaption effectively mitigates fine-grained hallucination in VLVMs, it remains subject to certain limitations. Firstly, the rewriting solely relies on keywords, such as verbs, nouns, and adjectives, disregarding structural information and the interrelations among the keywords. This limitation hinders the preservation of crucial details from the original image caption. Secondly, to train any LVLM, a small corpus of alignment-enriched rewrites is required, despite the minimal training costs involved.

\section{Ethics Statement}
To evaluate the proposed ReCaption, this paper presents a small sef to evaluation dataset FOHE, and we discuss some related ethical considerations here. 
(1) Intellectual property. FOHE is collected from the validation set of MSCOCO, which is licensed under Creative Commons Attribution 4.0. This license lets you build upon our evaluation.
(2) Treatments. We annotate the small evaluation data by our research group with agreed salaries and workloads.
(3) Controlling Potential Risks. The texts in FOHE do not involve private information and social issues.

\bibliography{anthology,custom}
\bibliographystyle{acl_natbib}

\appendix

\section{Appendix: More Examples}
\begin{figure*}[!htp]
  \centering
  \includegraphics[width=1.0\linewidth]{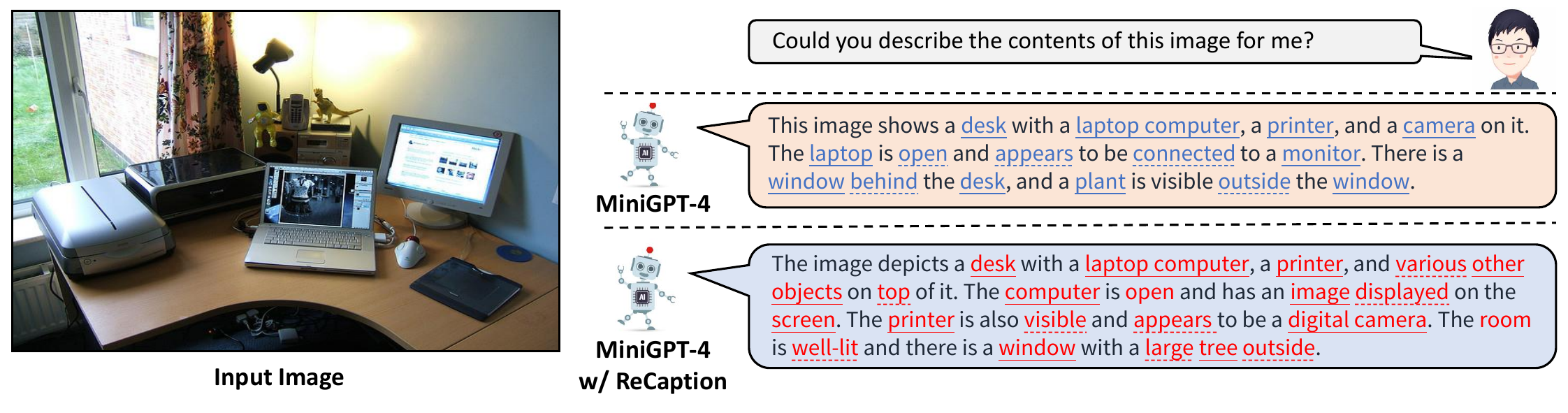}
  \caption{An illustrative example is presented to compare the output of MiniGPT-4 and MinitGPT-4 with ReCaption.
  }
  \label{fig:ap1}
\end{figure*}

\begin{figure*}[!htb]
  \centering
  \includegraphics[width=1.0\linewidth]{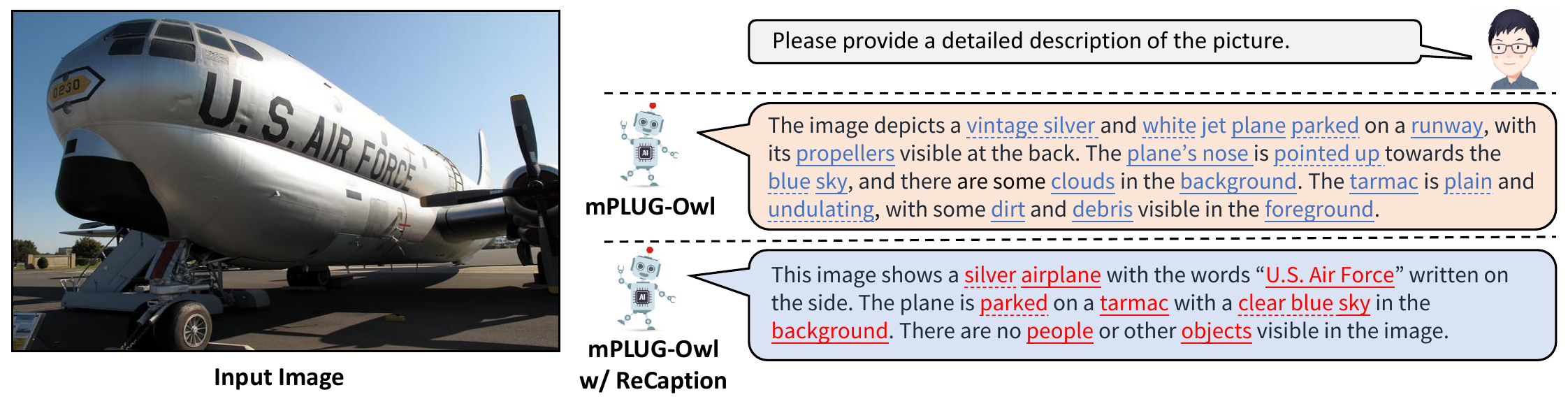}
  \caption{An illustrative example is presented to compare the output of mPLUG-Owl and mPLUG-Owl with ReCaption.
  }
  \label{fig:ap1}
\end{figure*}

\begin{figure*}[!htb]
  \centering
  \includegraphics[width=1.0\linewidth]{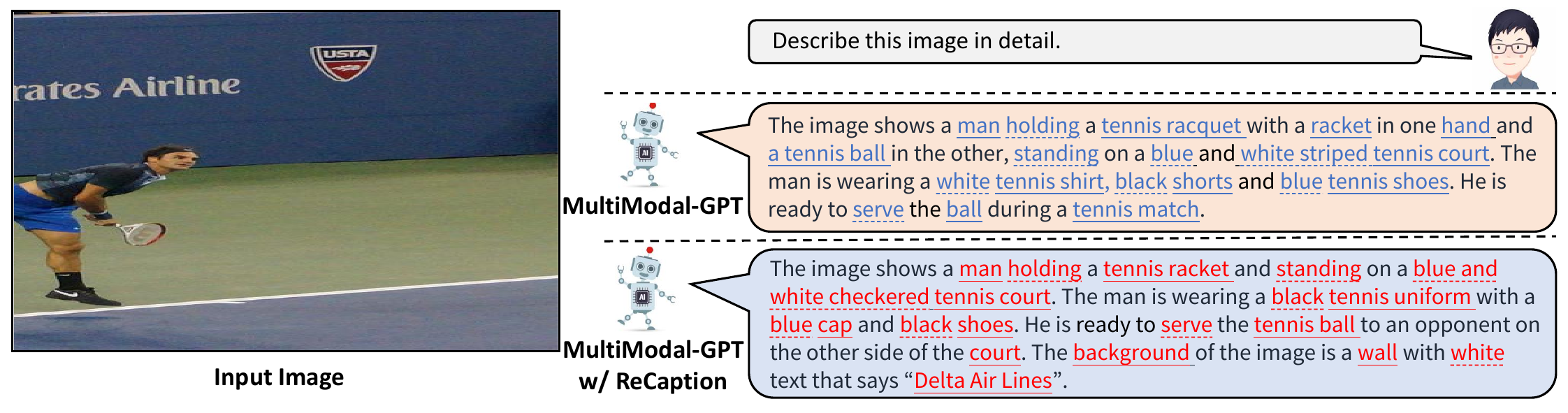}
  \caption{An illustrative example is presented to compare the output of MultiModal-GPT and MultiModal-GPT with ReCaption.
  }
  \label{fig:ap1}
\end{figure*}

\begin{figure*}[!htb]
  \centering
  \includegraphics[width=1.0\linewidth]{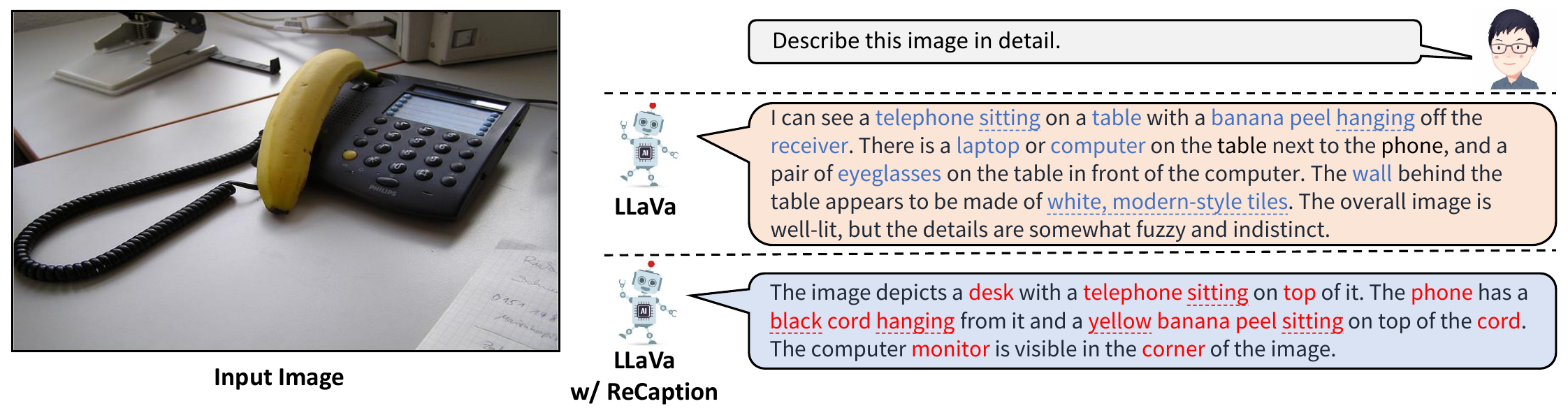}
  \caption{An illustrative example is presented to compare the output of LLaVa and LLaVa with ReCaption.
  }
  \label{fig:ap1}
\end{figure*}

\begin{figure*}[!htb]
  \centering
  \includegraphics[width=1.0\linewidth]{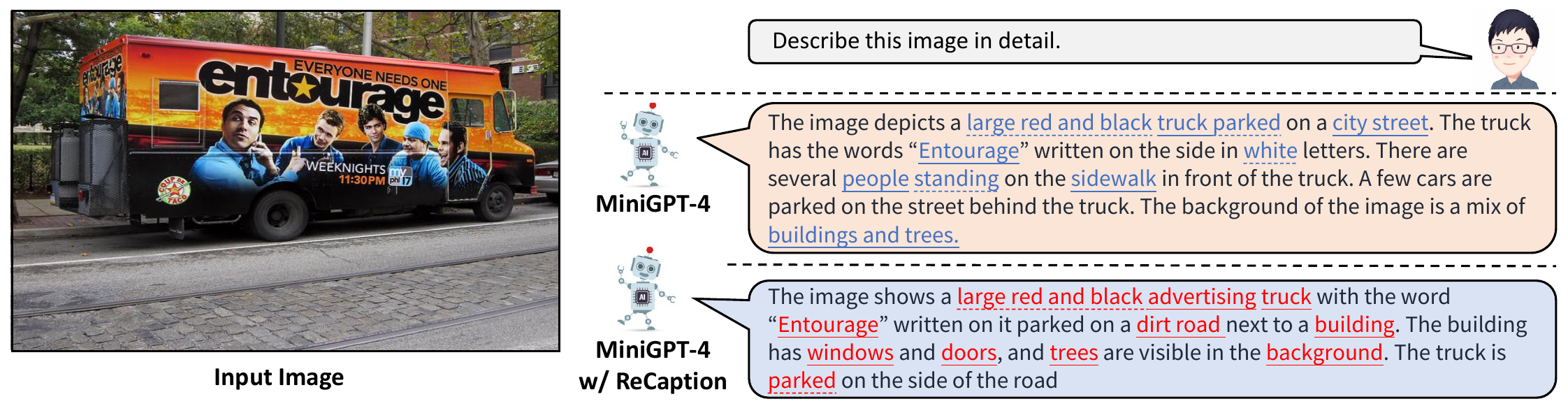}
  \caption{An illustrative example is presented to compare the output of MiniGPT-4 and MinitGPT-4 with ReCaption.
  }
  \label{fig:ap1}
\end{figure*}

\begin{figure*}[!htb]
  \centering
  \includegraphics[width=1.0\linewidth]{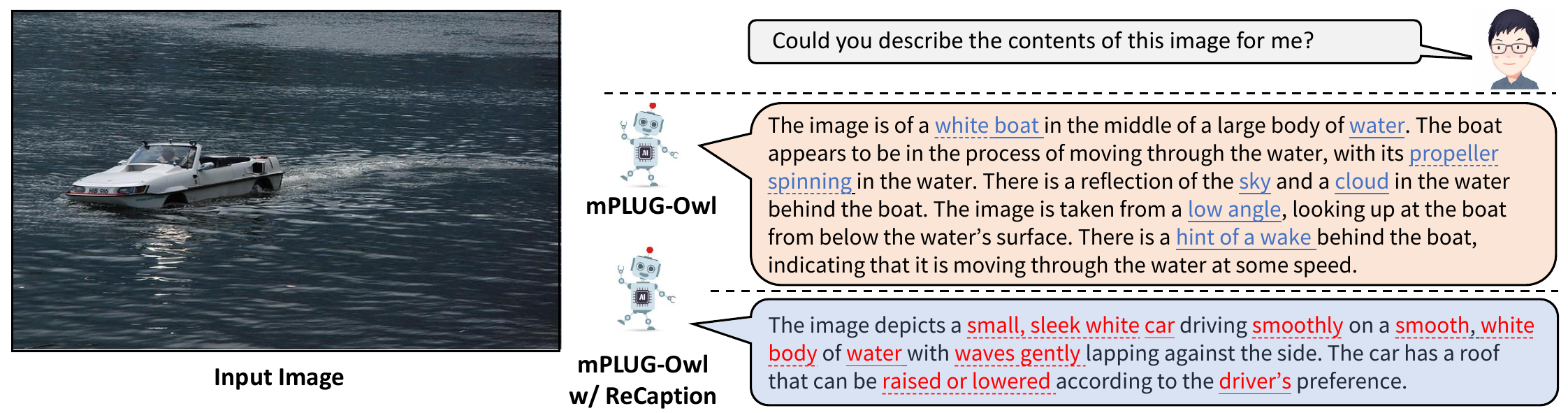}
  \caption{An illustrative example is presented to compare the output of mPLUG-Owl and mPLUG-Owl with ReCaption.
  }
  \label{fig:ap1}
\end{figure*}

\begin{figure*}[!htb]
  \centering
  \includegraphics[width=1.0\linewidth]{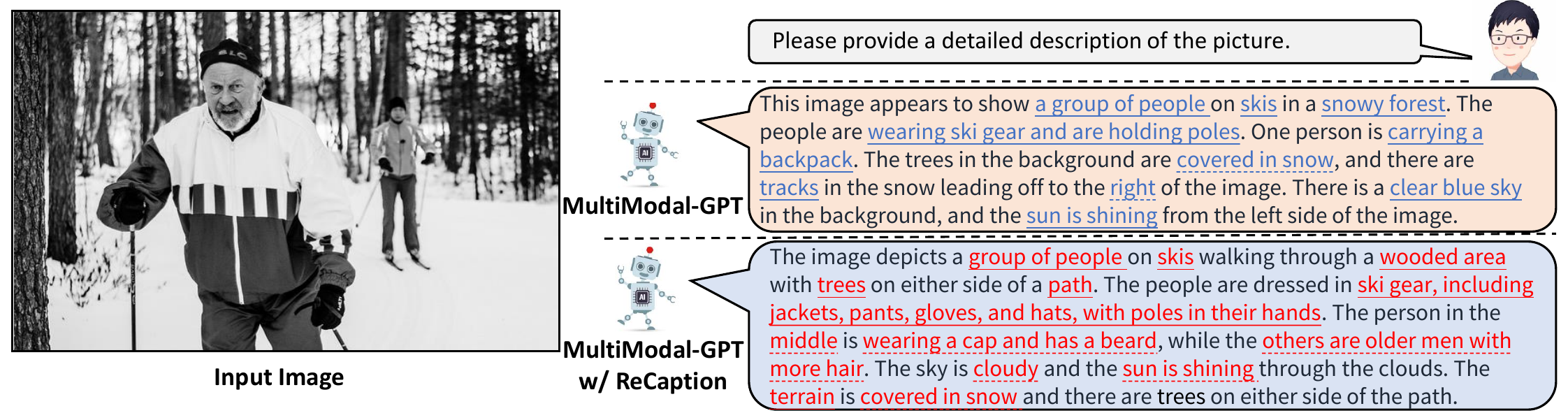}
  \caption{An illustrative example is presented to compare the output of MultiModal-GPT and MultiModal-GPT with ReCaption.
  }
  \label{fig:ap1}
\end{figure*}

\begin{figure*}[!htb]
  \centering
  \includegraphics[width=1.0\linewidth]{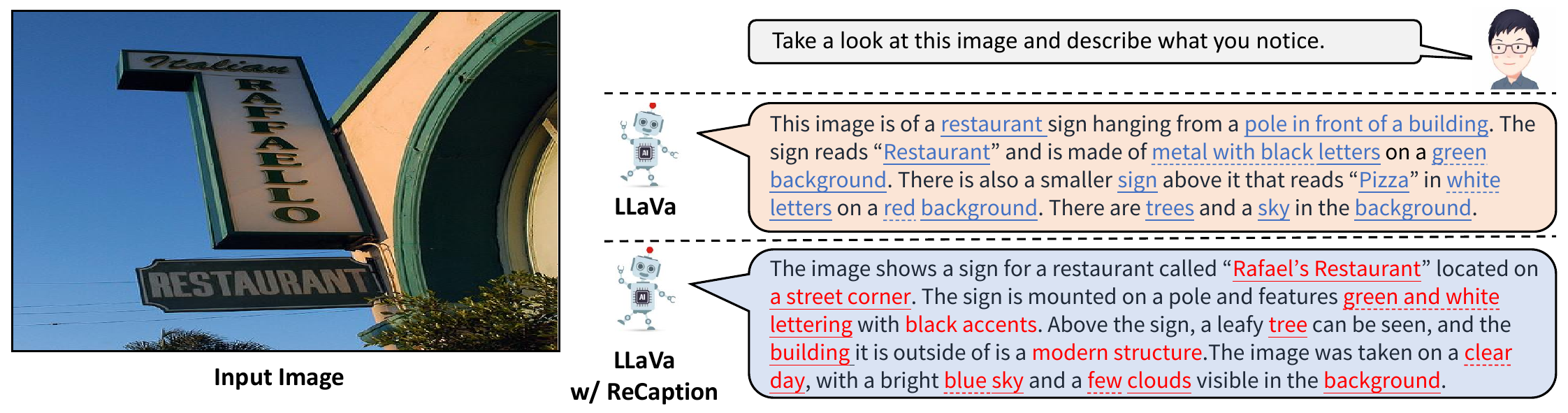}
  \caption{An illustrative example is presented to compare the output of LLaVa and LLaVa with ReCaption.
  }
  \label{fig:ap1}
\end{figure*}

\end{document}